\documentclass[sigconf, nonacm, screen]{acmart}

\AtBeginDocument{%
  \providecommand\BibTeX{{%
    \normalfont B\kern-0.5em{\scshape i\kern-0.25em b}\kern-0.8em\TeX}}}

\setcopyright{acmcopyright}
\copyrightyear{2021}
\acmYear{2021}
\acmDOI{}

\acmBooktitle{}
\acmPrice{}
\acmISBN{}

\usepackage{xcolor}
\usepackage{multirow}
\usepackage{hhline}

\begin{document}

\title{Aggregate Learning for Mixed Frequency Data}

\author{Takamichi Toda}
\email{toda_takamichi@cyberagent.co.jp}

\author{Daisuke Moriwaki}
\email{moriwaki_daisuke@cyberagent.co.jp}

\author{Kazuhiro Ota}
\email{ota\_kazuhiro@cyberagent.co.jp}

\affiliation{%
  \institution{CyberAgent, Inc.}
  \streetaddress{Shibuya}
  \city{Shibuya}
  \state{Tokyo}
  \country{Japan}
}

\renewcommand{\shortauthors}{Toda, Moriwaki and Ota}

\begin{abstract}
Large and acute economic shocks such as the 2007-2009 financial crisis and the current COVID-19 infections rapidly change the economic environment. In such a situation, the importance of real-time economic analysis using {\it alternative data} is emerging. Alternative data such as search query and location data are closer to real-time and richer than official statistics that are typically released once a month in an aggregated form. We take advantage of spatio-temporal granularity of alternative data and propose a \textbf{Mixed-Frequency Aggregate Learning (MF-AGL)} model that predicts economic indicators for the smaller areas in real-time. We apply the model for the real-world problem; prediction of the number of job applicants which is closely related to the unemployment rates. We find that the proposed model predicts (i) the regional heterogeneity of the labor market condition and (ii) the rapidly changing economic status. The model can be applied to various tasks, especially economic analysis.
\end{abstract}



\maketitle

\section{Introduction}


Devastating external economic shocks such as the 2007-2009 financial crisis and COVID-19 infections rapidly change economic circumstances. Amid these situations, real-time economic analysis is essential. The real-time analysis of the economic conditions is often called {\it Nowcasting} or {\it Economic Nowcasting}. Nowcasting provides important insights into the current economic status that are essential to appropriate policy responses including monetary and fiscal stimulus. An understanding of the on-going economic situation is also needed by private companies who make decisions on investment and employment. 

Nowcasting often takes advantage of {\it alternative data}, non-standard data such as search queries, location data, SNS data, and satellite images~\cite{varianPredictingPresentGoogle2009,damuriGoogleItForecasting2010,askitasGoogleEconometricsUnemployment2009,moriwakiNowcastingUnemploymentRates2020,indaco_twitter_2020,galimberti_forecasting_2020}. These data are suitable for economic nowcasting because of their high frequency. While most of the official statistics are reported once a month, alternative data is usually recorded in real-time. Another beauty of alternative data is its {\it granularity}. While official statistics usually report state-level statistics alternative data can provide detailed information up to the individual-level.

In response to the surging demand for alternative data, several works have already emerged. Tech giants such as Apple, Google, and Facebook provide mobility statistics based on their proprietary data~\cite{COVID19CommunityMobilityb,FacebookDataGood2020,COVID19Mobility}. These data help us understand several aspects of economic status. However, it is hard to understand the whole picture of the economy itself because the relationship between these alternative data-based statistics and familiar economic indices such as GDP and unemployment rates is unclear. 
To fill this gap, many nowcasting/forecasting models that use high-frequency data such as Google search query to predict economic variables are proposed~\cite{varianPredictingPresentGoogle2009,choiPredictingPresentGoogle2012,askitasGoogleEconometricsUnemployment2009,damuriGoogleItForecasting2010,suhoyQueryIndices2008,pavlicekNowcastingUnemploymentRates2015,moriwakiNowcastingUnemploymentRates2020}. Combining high-frequency data like Google search query and low-frequency data such as monthly unemployment rates and quarterly GDP has been actively studied by economists~\cite{ghyselsMIDASRegressionsFurther2007,ghysels_macroeconomics_2016}.

One caveat of these studies is that they do not fully extract the granularity of the alternative data. Target variables of these studies are often aggregated statistics such as GDP and unemployment rates. Although the predictor variables have a greater granularity, the predicted values are aggregated at the state or even national level. However, the governments, especially local governments need to deal with heterogeneity among small areas. For example, the bankruptcy of a large automaker will critically affect the local economy where the factories are located but does not affect the labor market at a national level. Authorities need to be sensitive to not only temporal changes but also regional heterogeneity.

The problem is that there is no granular data in the official statistics. That is, there is no label for the granular level. The problem is called {\it aggregate output learning} or {\it aggregate learning} and there has been only a few research papers has been published until recently~\cite{musicantSupervisedLearningTraining2007a}. However, the problem has been taking attention from an increasing number of researchers~\cite{dervalAggregateLearningApproach2020,lawVariationalLearningAggregate2018}.

We, here, combine mixed-frequency data literature in economics and aggregate learning literature in machine learning to fully utilize the richness of alternative data and provide an example of useful application. Our \textbf{Mixed-Frequency Aggregate Learning (MF-AGL) model} takes advantage of spatio-temporal granularity of alternative data and predicts economic indicators in smaller areas in real-time, which are not possible by standard forecasting models. The model also updates the prediction in real-time using high-frequency data without high-frequency target data. To train the model, we define a novel loss function for spatio-temporally aggregated label data and granular predicted values. More specifically, we aggregate predicted values for small areas and calculate loss using an aggregated level. We also calculate the loss for each high-frequency features so that the model can learn the mixed-frequency structure of the data. That means we reuse the same label data over and over.

We apply the model to the real-world problem; prediction of the number of job applicants for smaller areas in Japan. The number of job applicants reflects the condition of the labor market because job applicants are also unemployed persons. An acute increase in the number of job applicants implies deterioration of labor market conditions.  

To our best knowledge, the present work is the first to propose a novel method combining mixed frequency data and aggregate learning. We also demonstrate its practical importance in a real-world application. While we applied the model to the nowcasting of the labor market, the model can be applied to any task that contains (i) infrequent and aggregated indices such as GDP and (ii) spatio-temporally granular data.

In the following sections, we first discuss the background and related works in Section~\ref{sec:background and related work}, then state the problem setting and describe the proposed model in Section~\ref{sec:problem setting and method}. We then show the experiment results with the detail of the data and data pipeline in Section~\ref{sec:experiment}. Finally, Section \ref{sec:conclusion}concludes.

\section{Background and Related Works}
\label{sec:background and related work}
\subsection{Economic Nowcasting with Mixed-Frequency Data}
Nowcasting using mixed-frequency data has been an active research area~\cite{ghyselsMIDASRegressionsFurther2007,ghysels_macroeconomics_2016,uematsu_highdimensional_2019,mogliani_bayesian_2019,abrahamBigData21st2019,bai_state_2013,marcellino_factor-midas_2007}. In particular, nowcasting of labor market statistics with alternative data date back to the late 2000s~\cite{damuriGoogleItForecasting2010,varianPredictingPresentGoogle2009,askitasGoogleEconometricsUnemployment2009}. They suggested the potential predictive power of search query data. While most of the studies utilize web data represented by Google trends, Moriwaki (2020)~\cite{moriwakiNowcastingUnemploymentRates2020} uses smartphone GPS data to predict unemployment rates.

Unprecedented COVID-19 pandemic reminds the importance of economic nowcasting ~\cite{marketplaceWeNeedChange2020}. The government needs to recognize the rapidly-changing economic environment to take appropriate actions. In response to the surging demand, giant tech companies such as Google, Facebook, and Apple have been contributing by providing alternative data-based indices~\cite{COVID19CommunityMobilityb,FacebookDataGood2020,COVID19Mobility}.

Among them, mobility data has gained a lot of attention. Mobility data provides insight into how people change their behavior in life. Monitoring people's mobility patterns is an important example~\cite{kraemer_effect_2020,huang_understanding_2020,moriwaki_nudging_2020}. Mobility pattern reflects people's shopping behavior in physical stores, leisure, and travel.

In addition, more complicated economic activity can be assessed. One notable example is unemployment~\cite{moriwakiNowcastingUnemploymentRates2020}. In Japan, unemployed persons are required to visit public employment offices to collect unemployment insurance benefits. Hence the number of visitors is correlated to the number of unemployed persons. Official statistics only show monthly and prefecture-level data for the number of unemployment insurance benefits. The monthly statistics are usually released a month after the end of the month. Real-time data for this number is of high value in economic nowcasting.


\subsection{Aggregate Output Learning}
In their seminal work\cite{musicantSupervisedLearningTraining2007a}, Musicant et al.(2007) first defined the {\it aggregate output learning} problem in which the labels are only available in an aggregated form. They investigate various machine learning models that are applicable to the problem. The aggregate output learning problem has been recently re-investigated in various research~\cite{yousefiMultitaskLearningAggregated2019,lawVariationalLearningAggregate2018,dervalAggregateLearningApproach2020}.

The present work is also related to unsupervised learning for super-resolution~\cite{shocher2018zero} and video interpolation~\cite{reda_unsupervised_nodate} in the computer vision literature, which aims at recovering granularity from the data without labels. 


\section{Problem Setting and Method}
\label{sec:problem setting and method}
\subsection{Problem Setting}
\label{sec:problem setting}
Let $y_{t}^p \in \mathbb{R}$ be a {\it target} variable which is of interest of economists (e.g. unemployment rate and GDP). $p$ stands for some larger area such as nation and state and $t$ stands for some longer time period such as quarter and month. Let $x_\tau^q \in \mathbb{R}$ be {\it feature} variable which is correlated with the target variable $y_t^p$. $q$ stands for some smaller area such as city and county and $\tau$ stands for shorter time period such as day. The difference between $t$ and $\tau$, and  $p$ and $q$
 are the {\it granularity}. In particular, an area $p$ is divided by multiple small areas $q$s  and time period $t$ is divided by multiple short time period $\tau$. We use mapping function $\mu$ and $\pi$ to describe the relationship. In particular, $\mu(\tau) = t$ indicates that time $\tau$ belongs to $t$ and $\pi(q) = p$ indicates area $q$ belongs to $p$.

 Our goal is to find a predictor $f$ which predict $y^q_{t0}$  from granular data  $(x^q_\tau)_{\tau \le {\tau0}}$ , where $\mu(\tau_0) = t_0$, (i.e. $\tau_0$ belongs to $t_0$). Notice that the superscript is not larger area $p$ but small area $q$. That is, $f$ predicts  $y_t^{q}$ instead of $y^p_t$. 
 
Table~\ref{tab:sample_data} illustrates the data structure of our problem. Feature vector $\mathbf{x}$ is observed for City 1 and 2 and for each day. But output value $y$ is only observed for prefecture (assume the prefecture comprise of only two cities) and each month. We want to predict monthly values for each city (i.e., $y_{\rm Jan}^{1}$ and $y_{\rm Jan}^{2}$). Since feature vector $\mathbf{x}$ is collected in real-time, we want to update the prediction using the latest information. As shown in the Table~\ref{tab:sample_data} the predicted values are changing according to the feature vector. That is, $\hat{y}_{\rm Jan}|\mathbf{x}_{\rm Jan1} \neq \hat{y}_{\rm Jan}|\mathbf{x}_{\rm Jan2} $.
In this way, we can fully utilize the real-time and granular data for the forecast.

\begin{table}[htbp] 
    \centering
    \caption{An illustration of output data $y$, Input data $\mathbf{x}$, and predicted data $\hat{y}$. The predicted data is updated every day using latest information although the label data is only available for spatially and temporally aggregated value.} 
    \begin{tabular}{|l|c|c|c|c|c|} \hline
   \multicolumn{1}{|c}{}  & \multicolumn{1}{|c|}{output} & \multicolumn{2}{|c|}{input} & \multicolumn{2}{c|}{pred} \\ \hline
     & \multirow{2}{*}{Pref}  & \multicolumn{2}{|c|}{Pref.A} & \multicolumn{2}{c|}{Pref.A} \\ \hhline{~~----}
     &  & City1 & City2 & City1 & City2  \\ \hline
        Jan 1 & \multirow{4}{*}{$y_{\rm Jan}$} & $\mathbf{x}_{\rm Jan 1}^1$ & $\mathbf{x}_{\rm Jan 1}^1$ & $\hat{y}_{\rm Jan}^1 \mid \mathbf{x}_{\rm Jan 1}^1$ & $\hat{y}_{\rm Jan}^2 \mid \mathbf{x}_{\rm Jan 1}^2$   \\ \hhline{-~----} 
        Jan 2 &  & $\mathbf{x}_{\rm Jan 2}^1$ & $\mathbf{x}_{\rm Jan 2}^1$ & $\hat{y}_{\rm Jan}^1 |\mathbf{x}_{\rm Jan 2}^1$ & $\hat{y}_{\rm Jan}^2 \mid \mathbf{x}_{\rm Jan 2}^2$   \\ \hhline{-~----}
        $\cdots$ & & $\cdots$ & $\cdots$ & $\cdots$ & $\cdots$ \\ \hhline{-~----}
        Jan 31 & & $\mathbf{x}_{\rm Jan 31}^1$ & $\mathbf{x}_{\rm Jan 31}^2$ & $\hat{y}_{\rm Jan}^1 \mid \mathbf{x}_{\rm Jan 31}^1$ & $\hat{y}_{\rm Jan}^2 \mid \mathbf{x}_{\rm Jan 31}^2$ \\ \hline
        Feb 1 & $y_{\rm Feb}$ & $\mathbf{x}_{\rm Feb 1}^1$ &$\mathbf{x}_{\rm Feb 1}^2$ & $\hat{y}_{\rm Feb}^1 \mid \mathbf{x}_{\rm Feb 1}^1$ & $\hat{y}_{\rm Feb}^2 \mid \mathbf{x}_{\rm Feb 1}^2$  \\ \hline
    \end{tabular}
    \label{tab:sample_data}
\end{table}

While we conduct spatial disaggregation, we directly predict the aggregated value for a longer period. That is, we predict $y^q_t$ rather than $y_\tau^q$. This seems simple supervised learning but it is not. To see this, let $\tau$ be Jan 15, 2020. Then $t$ is Jan 2020. We have data only for the past. That is, we have $\mathbf{x}_{\rm Jan 1}^q, \mathbf{x}_{\rm Jan 2}^q, \cdots , \mathbf{x}^q_{\rm Jan15}$ but do not have ${\mathbf{x}_{\rm Jan16}^q, \mathbf{x}_{\rm Jan 17}^q, \cdots, \mathbf{x}_{\rm Jan 31}^q}$. Nowcasting predicts the aggregated value using incomplete time series data~\cite{ghyselsMIDASRegressionsFurther2007}. For economists, the monthly value $y_t^q$ is more informative than the daily value $y_\tau^q$. $y_\tau^q$ represents the economic status in a very short term while $y_t^q$ represents the forecast of the longer time period. For example, let $y_t^q$ be the unemployment rate for January 2020 and $y_\tau^q$ be that for January 1, 2020. While the daily movement of the unemployment rate is interesting for stock traders who are eager to know the short-term fluctuation, the forecast of the monthly unemployment rate is of prime interest for policymakers who need to know the trend of economic status. Furthermore, policy interventions take some time to be implemented.

On the other hand, geographically granular estimates $y^q_t$  are much more useful than aggregated value especially for local governments who need precise information about their local economy. Nevertheless, our method is easy to transform into a daily version.  In sum, our goal is to find a good predictor $f(\mathbf{x}_\tau^q)$ for $y^q_t$, $\mu(\tau) = t$.


\subsection{Aggregate Learning for Mixed-Frequency Data}
In this section, we describe our proposed model.
\subsubsection{Aggregate Function}
Following the aggregate learning literature\cite{musicantSupervisedLearningTraining2007a,dervalAggregateLearningApproach2020}, we first define the aggregation function as,
\begin{eqnarray}
\label{eq:aggregate function}
\hat{y}_t^p = \sum\limits_{q \in p}\omega_q f(\mathbf{x}_\tau^q, \boldsymbol{\phi}_\tau^q),
\end{eqnarray} 
where $\omega_q \in [0, 1]$ is weight. The weight controls for the share of the values for each granular area in the large area. In the simplest case (including the application discussed below), $\omega_q$ is set to one. In these cases, $f$ can learn the actual values of the target from the features. However, when we only obtain {\it normalized} features such as population per acre or averaged age, we need weighted sum. In these cases, the weights are pre-determined based on real data and knowledge. Area and population are often surveyed at a fine granularity by censuses and can be used as good proxies of the share.

\subsubsection{Mixed -Frequency Aggregate Learning Model}
Predictor $f$ predicts the outcome values for small areas. In contrast to standard supervised learning, true values for the predictor can not be observed. In another word, the predictor predicts {\it latent} values. The main feature vector $\mathbf{x}_\tau^q$ is of granular information such as search queries, posts in social networking service (SNS), point of sales (POS), credit card, and mobility data. These data are often called {\it alternative data}. The vector $\mathbf{x}_\tau^q$ contains the current value $x_\tau^q$ and its lagged values. That is, $\mathbf{x}_\tau^q = (x_\tau^q, x_{\tau-1}^q, x_{\tau-2}^q,\cdots)$. 
Forecasters want to update the prediction when new data arrive. That is, $\tau$ can be any timing. For example, Let $t$ be April 2020 and $\tau_1, \tau_2, \cdots, \tau_{30}$ be April 1st,  2nd $,\cdots,$ 30th. Then, $\mathbf{x}_{\tau1}$ and $\mathbf{x}_{\tau2}$ should give different prediction. However, the label data are only available for complete data $\mathbf{x}_{\tau30}^q = (x_{\tau30}^q, x_{\tau29}^q,\cdots, x_{\tau1}^q)$. $x_{\tau30}^q$ is missing in $\mathbf{x}_{\tau29}^q$. In economics, missing data is dealt with by either (i) training model on data with the same missing structure or (ii) imputation of missing data. In the above example, a forecaster who adopts (i) uses only data from the 1st to 29th day for each month to keep the same missing structure when training their model. Naturally speaking, it causes a huge loss of information. A forecaster who resorts to (ii) needs to prepare another model to conduct imputation. 

In our model, the missing-ness is treated by introducing auxiliary feature vector $\boldsymbol{\phi}$ and non-linearity in the parameters. The vector $\boldsymbol{\phi}_\tau^q$ contains one-hot encoded year, month, day, larger area $p$, and small area $q$. By doing so, predictor $f$ can learn the missing structure of the data and appropriately use the information. 
The non-linearity of the parameter is essential in this sense. We adopt a simple recurrent neural network (RNN) to make the predictor flexible (Fig.~\ref{fig:lstm_arch}). As a result, we can fully utilize the information and make it end-to-end.

\subsubsection{Model Training}

The predictor is trained by minimizing the MSE loss,
\begin{eqnarray}
\label{eq:loss}
\mathcal{L}(f) = \sum\limits_p \sum\limits_t \sum\limits_{\tau: \mu(\tau) = t} \left( y_t^p - \sum\limits_{q:\pi(q) = p}\omega_q f(\mathbf{x}_\tau^q, \boldsymbol{\phi}_\tau^q) \right)^2.
\end{eqnarray}

At each data point $(\tau, p)$, the predictor $f$ predicts the values for small area $q$ at short term period $\tau$. The predicted values are weighted-average and the loss is calculated.   

\subsubsection{Prediction}
Trained predictor $f$ is used for the prediction using granular data. Forecaster can use real-time data $\mathbf{x}_\tau^q$ to predict the smaller area's current status as $\hat{y}^q_t = f(\mathbf{x}_\tau^q)$. The good news is that the forecaster does not need to re-train the model until new label data $y$ is released. She can re-use the same model for an extended period.

\section{Experiment with Japanese Job Application Data}
\label{sec:experiment}
We apply the MF-AGL model to analyze the real-world economy. In particular, we apply the model to predict the number of job applicants in Japan. The Japanese government releases official statistics on the number of persons who file job applications to public employment offices on monthly basis. In Japan, unemployed persons need to file a job application to the public employment office to take up the unemployment insurance benefits. The number of job applicants is counted at 544 public employment offices, the official statistics summarizes the number for 47 prefectures. The number of job applications is a good proxy for the number of unemployed persons. This is very similar to the unemployment insurance claims statistics in the U.S., which is considered one of the most important economic indicators by economists.

The reasons we chose the problem are the following. First, unemployment is a huge tragedy; It leads to loss of income and also loss of contact with society, which causes economic and mental hardship. Real-time analysis of labor market conditions is essential for a swift policy response. Second, the prefecture-level data provided by the reports are too rough for the appropriate policy response. In Japan, each prefecture has a population of several million to ten million. More granular statistics are needed for careful policy intervention. Third, monthly updates of the reports are too infrequent. Amid the COVID-19, the deterioration of the market is very fast. Looking at monthly data for one or two months ago is not very meaningful. We need a real-time update for the data.

\begin{figure}[tbhp]
 \centering
 \includegraphics[width=0.7\linewidth]{"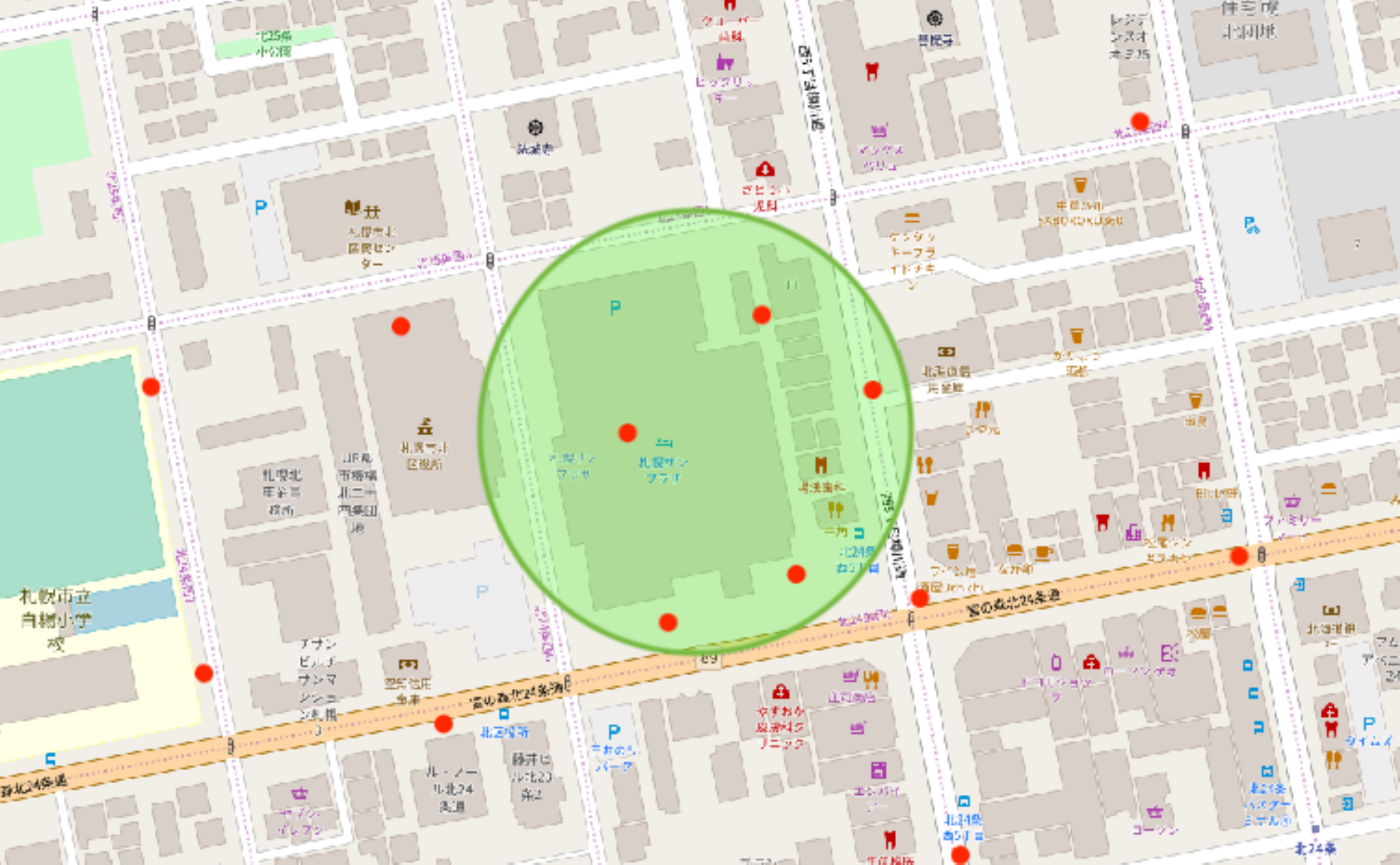"}
 \caption{GPS readings around a public employment office. The points are artificially generalized and does not represent real human location history.} 
 \label{fig:hellowork}
\end{figure}

Fortunately, there is a good {\it alternative data} for the number of job applicants. As Moriwaki~\cite{moriwakiNowcastingUnemploymentRates2020} shows GPS data from smartphones has good predictive power for the number of unemployed persons. In this study, we utilize similar datasets. As shown in Fig.~\ref{fig:hellowork}, the GPS readings around public employment offices indicate the visits to the office. In contrast to \cite{moriwakiNowcastingUnemploymentRates2020} who only count the number of GPS readings inside the radius of the offices, we extract rich features from these data. The detail of the feature extraction is provided in Section~\ref{sec:feature extraction from GPS readings}

\begin{figure*}[tbhp]
 \centering
 \includegraphics[width = 0.7\linewidth]{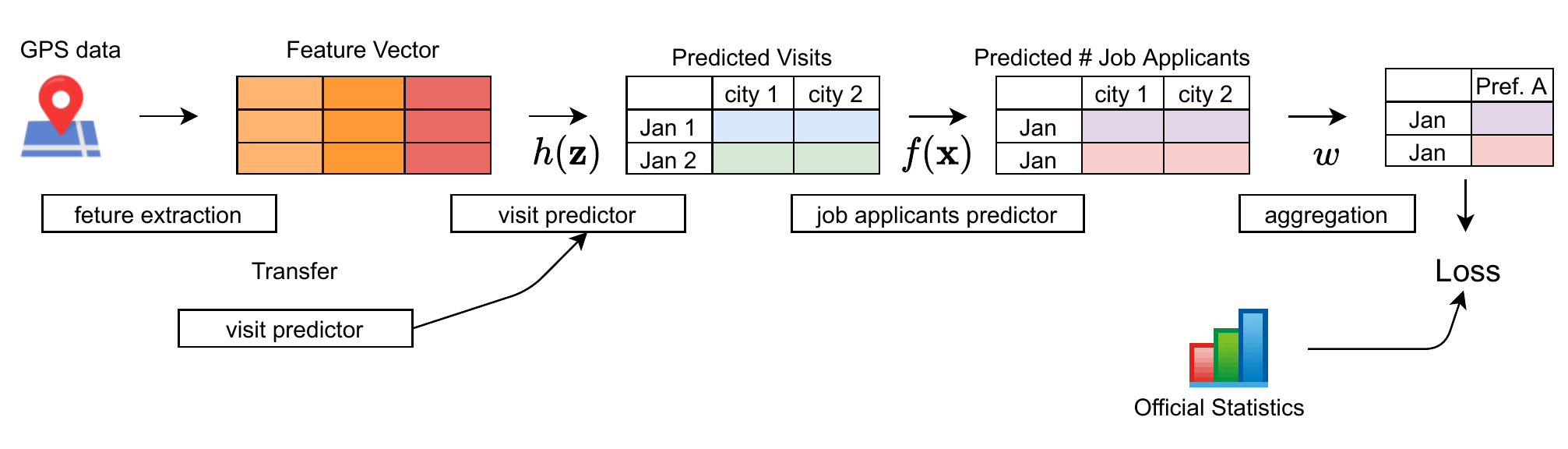}
 \caption{Overview of the data pipeline. } 
 \label{fig:data pipeline}
\end{figure*}

The whole process is summarized in Fig.~\ref{fig:data pipeline} We first extract feature vector $\mathbf{z}$ from raw GPS readings taken from mobile apps. Then the visit predictor predicts $\mathbf{x_\tau^q}$, the number of visits to public unemployment offices located in each city.  Then another predictor $f$ predicts $y_t^q$, the number of job applicants for each office.  The visit predictor is trained on the different domain and transferred to the task. The transfer learning is explained in Section~\ref{sec:visits prediction using transfer learning}. 

\subsection{Datasets}
We use three datasets as follows.
\subsubsection{Reports on Employment Service by Ministry of Health, Labour, and Welfare}
The reports on employment service is a monthly official statistics released by the Ministry of Health, Labour and Welfare (MHLW). The reports contain the monthly number of job applicants by prefecture. The data is publicly available on the webpage of MHLW~\cite{ministry_of_health_welfare_and_labour_reports_nodate}.

\subsubsection{GPS data from Smartphones}
GPS data is taken from various mobile apps from Jan 1, 2016, to Oct 31, 2020. The data is anonymized and not related to privacy. The number of users ranges from several hundred thousand to several million. The data contains tuples (latitude, longitude, timestamp). The data contains no private information.

\subsubsection{Locations of Public Employment Offices}
The public employment offices are established by the Japanese government based on the ILO treaty. The location of the offices is publicly available on the webpage of the MHLW. We use the location with the GPS trajectories to extract feature values~\cite{ministry_of_health_welfare_and_labour_location_nodate}.


\subsection{Feature Extraction from GPS Readings}
\label{sec:feature extraction from GPS readings}
The crucial challenge is to accurately count visitors from noisy GPS readings. GPS is very noisy especially for the smartphone because the logs are usually very sparse. For example, many apps log the GPS records when the phone is moved. This algorithm aims at minimizing buttery consumption. As a result, the records are not recorded at the same intervals. Also, the accuracy of location deteriorates inside buildings as the signals from satellites are not reached. One possible solution is to rely on the machine learning techniques that denoise the data and extract visited POIs (Point-of-Interests). There are various approaches to do this job~\cite{gongDerivingPersonalTrip2014,keles_extracting_2017,nishida_extracting_2017,nishida_probabilistic_2014}. We find that the visited point extraction methods typically rely on (i) the number of stay points, (ii) the stop location, (iii) stop duration, and (iv) speed. We follow their approach and extract extensively rich information from location data to achieve high performance. The list of the features is presented in Table \ref{tab:geo-features}. 


The visit predictor in Fig.~\ref{fig:data pipeline} uses location trajectory to classify visit/non-visit to some POIs. 

\begin{table}[htpb]\small
    \centering
        \caption{List of Geo-Features. These features are extracted from raw GPS trajectories.}
    \begin{tabular}{p{0.35\linewidth}p{0.5\linewidth}} \toprule
    \multicolumn{1}{c}{feature} & \multicolumn{1}{c}{description} \\ \midrule
         \# records inside $X$ km & \# GPS records inside X kirometer radius around POI. $X$: 0.5-0.003. \\
         \# records outside 0.5km & \# GPS records outside 0.5 km radius around POI \\
         size & size of POI \\
         \# records inside building & \# GPS records inside POI \\
         mean speed & Averaged speed of the user \\
         max speed & Max speed of the user \\
         stay count & \# of stay points\\
         speed at 9 points & Speed at 9 nearest points from POI\\
         cosine at 9 points & Cosine extracted from the trajectory of users \\ \bottomrule
    \end{tabular}
    \label{tab:geo-features}
\end{table}

\subsection{Visits Prediction using Transfer Learning}
\label{sec:visits prediction using transfer learning}
\subsubsection{Transfer of Visit Predictor} 
Another challenge for visit prediction is that there is no {\it true} label for public employment office visitors. We transfer the visit predictor trained on the different proprietary source and transfer to the visit prediction for the main task; visits to public employment offices.

The visits predictor uses the extracted features as in Section~\ref{sec:feature extraction from GPS readings}. The predictor is implemented using LightGBM~\cite{keLightGBMHighlyEfficient2017}. The hyperparameter tuning is done with the LightGBM tuner. These features are powerful. The visit predictor achieved an AUC of 0.86 for the classification task for the original task (train: test = 9,091 : 4,479). 

\subsubsection{Visits Prediction for Public Employment Offices} Unemployed persons visit nearest public employment office to file an unemployment insurance claim.  There are 544 offices in Japan. We divide the entire country into 544 regions based on the location of the offices assuming each region is covered by the nearest office. We use the transferred predictor to predict the visit count to each office. 


\subsection{Job Applicants Predictor and Aggregation Function}
\label{sec:Job Applicants Predictor and Aggregation Function}
With the predicted visits count, we train the job applicants predictor that predicts the number of job applicants for each public employment office. Input to the model is described in Table \ref{tab:input variables}. The predictor uses past visit count and dummy variables extracted from the date of making a prediction. 

\begin{table}[htpb]\small
    \centering
        \caption{List of Input Variables in Job Applicants Predictor}
    \begin{tabular}{ll} \toprule
    \multicolumn{1}{c}{feature} & \multicolumn{1}{c}{description} \\ \midrule 
     visit counts & Predicted daily visit counts for past 31 days.   \\
     year dummy & One-hot encoded year dummy\\ 
     month dummy & One-hot encoded month dummy  \\
     day dummy & One-hot encoded day dummy  \\
     prefecture dummy & One-hot encoded prefecture dummy \\
     public office dummy & One-hot encoded office dummy \\
 \bottomrule
    \end{tabular}
    \label{tab:input variables}
\end{table}

\begin{figure*}[t]
    \centering
    \includegraphics[width=0.6\linewidth]{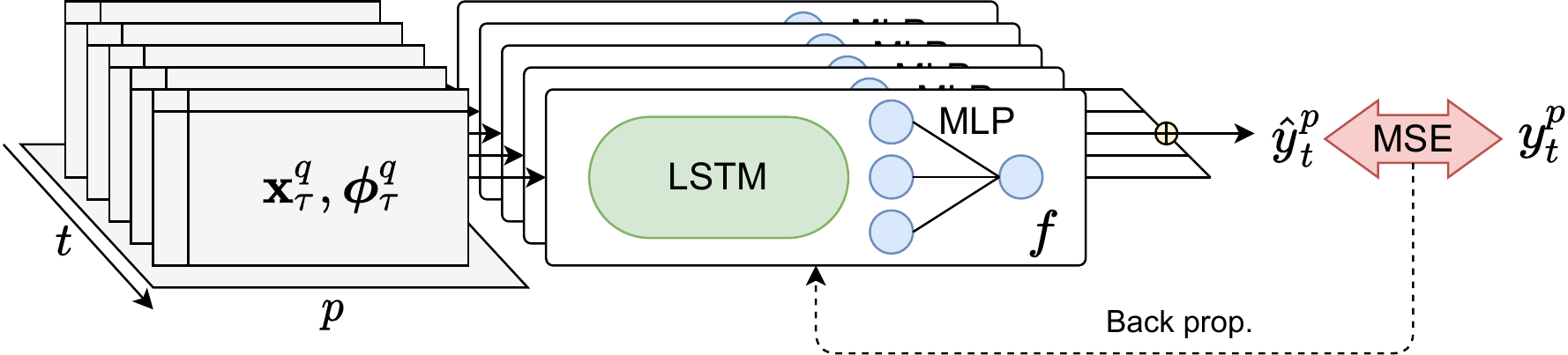}
    \caption{Our aggregate learning architecture for job applicants prediction. We use a simple recurrent neural network composed of an LSTM layer and a multi-layer perceptron as a predictor. Each daily prediction is aggregated as monthly visits count of large area. The loss value is computed by the MSE criterion on the training phase.}
    \label{fig:lstm_arch}
\end{figure*}

The training/prediction model is a simple recurrent neural network, and Fig.~\ref{fig:lstm_arch} shows the network architecture. This model uses an LSTM layer and a multi-layer perceptron to predict a daily visit count of small areas, and each prediction is aggregated to predict a monthly visit count of large areas according to Eq.~\eqref{eq:aggregate function}. We use Adam~\cite{kingma2015adam} solver for optimization with $\beta1 = 0.9, \beta2 = 0.999$, initial learning rate = 0.0001, no weight decay and no learning rate decay. We train our models for a total of 600 epochs with a batch size of 1 and the MSE loss described in Eq.~\eqref{eq:loss} on Tesla V100 GPU. We implemented the whole network in PyTorch~\cite{pytorch}.

\subsection{Spatial Disaggregation using MF-AGL model}
\label{sec:spatial disaggregation}
\begin{figure*}[tbhp]
 \centering
 \includegraphics[clip, width = .6\linewidth]{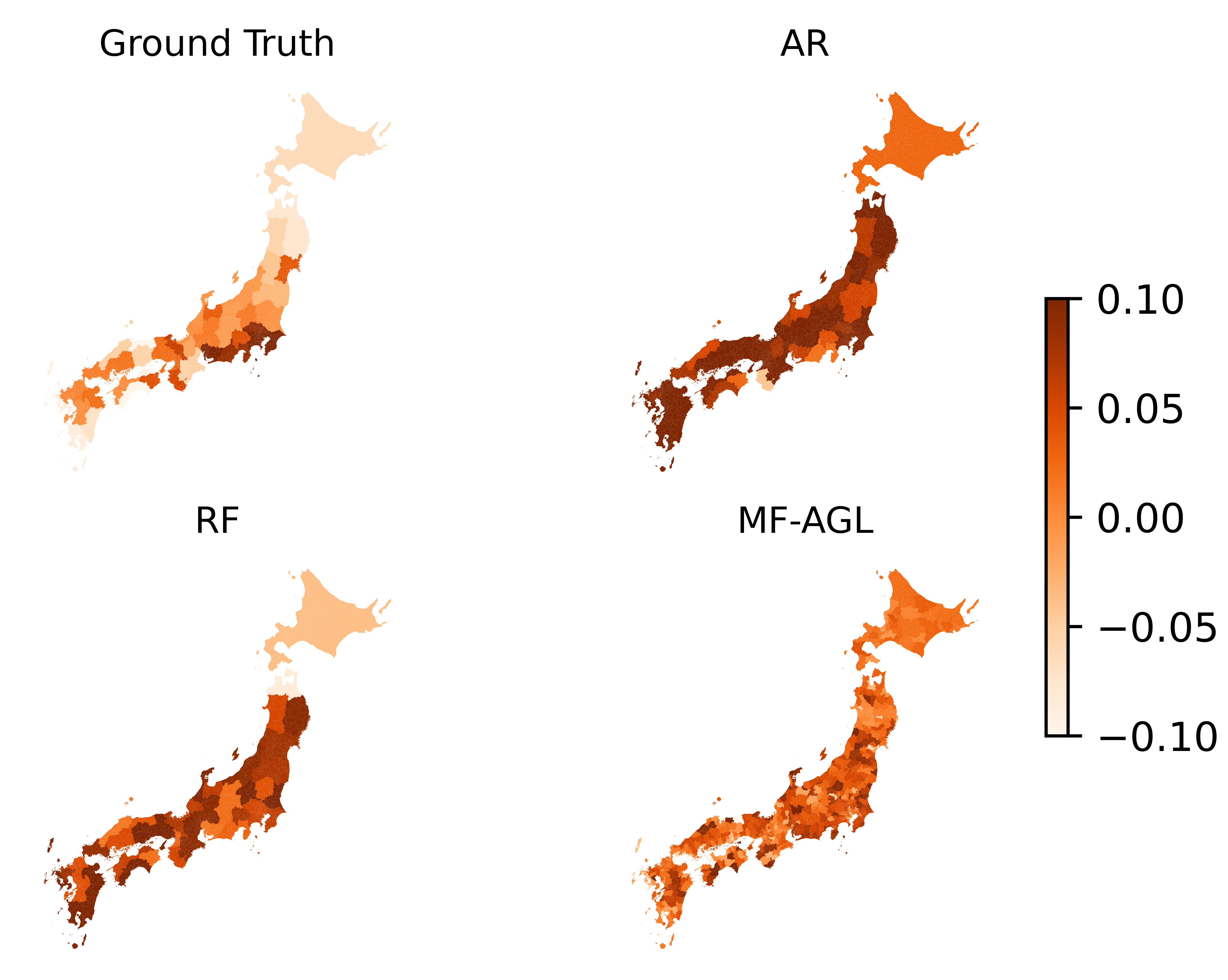}
 \caption{Year-over-year change in the number of job applicants in Japan. Due to the limitation of the space, we exclude Okinawa islands from the map. The color represents good (lighter)-bad (darker) conditions of labor market. \textbf{Ground Truth} shows the actual data from official statistics;\textbf{AR} shows the prediction made by the AR model with 11 lags;\textbf{RF} shows the predictions made by Random Forest model with 11 lags and year, month and prefecture dummy as input;\textbf{MF-AGL} shows the prediction made by MF-AGL model. MF-AGL model shows the much granular predictions.} 
 \label{fig:comparison}
\end{figure*}

Now we demonstrate the usefulness of the proposed model. Fig.~\ref{fig:comparison} shows the four maps represent regional heterogeneity of the change of the number of job applicants. The color represents decrease (good, bright) and increases (bad, dark) in the number of job applicants from the previous year (i.e., year over year).

The maps are generated by data from the actual Reports on Employment Service for October 2020 (Ground Truth), predictions by the proposed model (MF-AGL), predictions by an auto-regressive model (AR), and predictions by a Random Forest model (RF). The  auto-regressive (AR) model is a standard model for economic forecasting and time-series prediction. Random Forest (RF) is a standard machine learning method which is especially good for small sample. The AR model uses the number of job applications of past 11 months (i.e., 11 lags) as inputs, i.e., $\hat{y}^p_t = f^{AR}(y^p_{t-1}, \cdots, y^p_{t-11})$. RF model uses dummy variables for year, month, and prefecture as well as the number of job applications of past 11 months (i.e., 11 lags).

The MF-AGL model is trained on the data from October 2016 to September 2020. Then the model use feature extracted on October 31, 2020. The AR and RF model is trained on the data from October 2016 to September 2020 and use the data from September 2020 as input. It seems unfair that only the MF-AGL model uses the data from October 2020. However, this is the strength of the nowcasting model.

To see this, Table~\ref{tab:schedule} shows the schedule of data availability. On October 1, 2020, we only have label data for July 2020 and before because the Reports for August 2020 are released on October 2. On the other hand, we have features $\mathbf{x}_{\rm Oct 1}^q$ in real-time. As such, on October 31, 2020, our MF-AGL model can use the latest values of features while the traditional AR and RF model can only use the label data for the last month.
\begin{table*}[phtb]\small
    \centering
    \caption{The schedule of data availability}
    \begin{tabular}{|l|c|c|c|c|}\hline
     & Oct 1 & Oct 2, $\cdots$,  Oct 29 & Oct30 & Oct31 \\ \hline
 output & $y_0^p, \cdots, y_{\rm July}^p$     & $y_0^p, \cdots, y_{\rm Aug}^p$ & \multicolumn{2}{c|}{$y_0^p, \cdots, y_{\rm Sep}^p$} \\ \hline   
      input &    $\mathbf{x}_{\rm Oct 1}^q$ & $\mathbf{x}_{\rm Oct 2}^q, \cdots, \mathbf{x}_{\rm Oct 29}^q$  & $\mathbf{x}_{\rm Oct 30}^q$  & $\mathbf{x}_{\rm Oct 31}^q$ \\ \hline 
 AR/RF & $\hat{y}_{\rm Oct}^p \mid y_0^p, \cdots, y_{\rm July}^p$ & $\hat{y}_{\rm Oct}^p \mid y_0^p, \cdots, y_{\rm Aug}^p$ & \multicolumn{2}{c|}{$\hat{y}_{\rm Oct}^p \mid y_0^p, \cdots, y_{\rm Sep}^p$} \\ \hline 
 MF-AGL & $\hat{y}_{\rm Oct}^q \mid \mathbf{x}_{\rm Oct1}^q$ & $\hat{y}_{\rm Oct}^q \mid \mathbf{x}_{\rm Oct2}^q,  \cdots, \hat{y}_{\rm Oct}^q \mid \mathbf{x}_{\rm Oct29}^q$ & $\hat{y}_{\rm Oct}^q \mid \mathbf{x}_{\rm Oct30}^q$ & $\hat{y}_{\rm Oct}^q \mid \mathbf{x}_{\rm Oct31}^q$\\ \hline
    \end{tabular}
    \label{tab:schedule}
\end{table*}
Now let's turn to the maps in Fig.~\ref{fig:comparison}. One significant difference of the MF-AGL model is its geographical granularity. While the other three figures only tell that the south-east areas are in bad condition (darker), the MF-AGL model shows that there is a mix of bad and good conditions at a granular level. We can see there are darker areas in the granular map that are bright in the ground truth data. The local governments need to take care of these hidden problems.

The other finding is that the AR and RF models are not good at prediction. To see this, we first aggregate the prediction by MF-AGL model at prefecture-level and calculate Mean Absolute Percentage Error (MAPE), a standard metric to evaluate forecasting models. 
\begin{eqnarray} \label{eq:mape}
{\rm MAPE} = \frac{1}{\mid \mathcal{P} \mid}\sum\limits_{p \in \mathcal{P}} \frac{ \mid y^p_t - \hat{y}_t^p \mid }{y_t^p}
\end{eqnarray}
The reason we use MAPE is that the number of job applications in each prefecture is proportional to the population. Other metrics such as mean squared error (MSE) and mean absolute error (MAE) are more affected by an error in the populous prefecture while MAPE treats each prefecture equally.

The results are shown in Table~\ref{tab:MAPE}. The MF-AGL obtained the lowest error. Although the prediction performance at the aggregate level is not the priority of the MF-AGL model, this result highlights the robustness of the prediction of the model.
\begin{table}[htpb]
    \centering \small
    \caption{Mean Percentage Errors by models}
    \begin{tabular}{lccc} \toprule
model & mape (\%) & Std. Err & N\\ \midrule
AR & 13.41 & (0.66) & 47\\
RF & 8.11 & (0.67) & 47\\ 
MF-AGL & \textbf{7.78} & (1.03) & 47\\ \bottomrule
\end{tabular} 
\begin{minipage}{.7\linewidth} 
{\small {\it Note}: The mean percentage error by the model calculated by eq.~\ref{eq:mape}. . The values are calculated for the prediction shown in Fig.~\ref{fig:comparison}.}
\end{minipage}
    \label{tab:MAPE}
\end{table}
\subsection{Real-time Labor Market Analysis using MF-AGL Model}
\begin{figure*}[tbhp]
 \centering
 \includegraphics[clip, width = .8\linewidth]{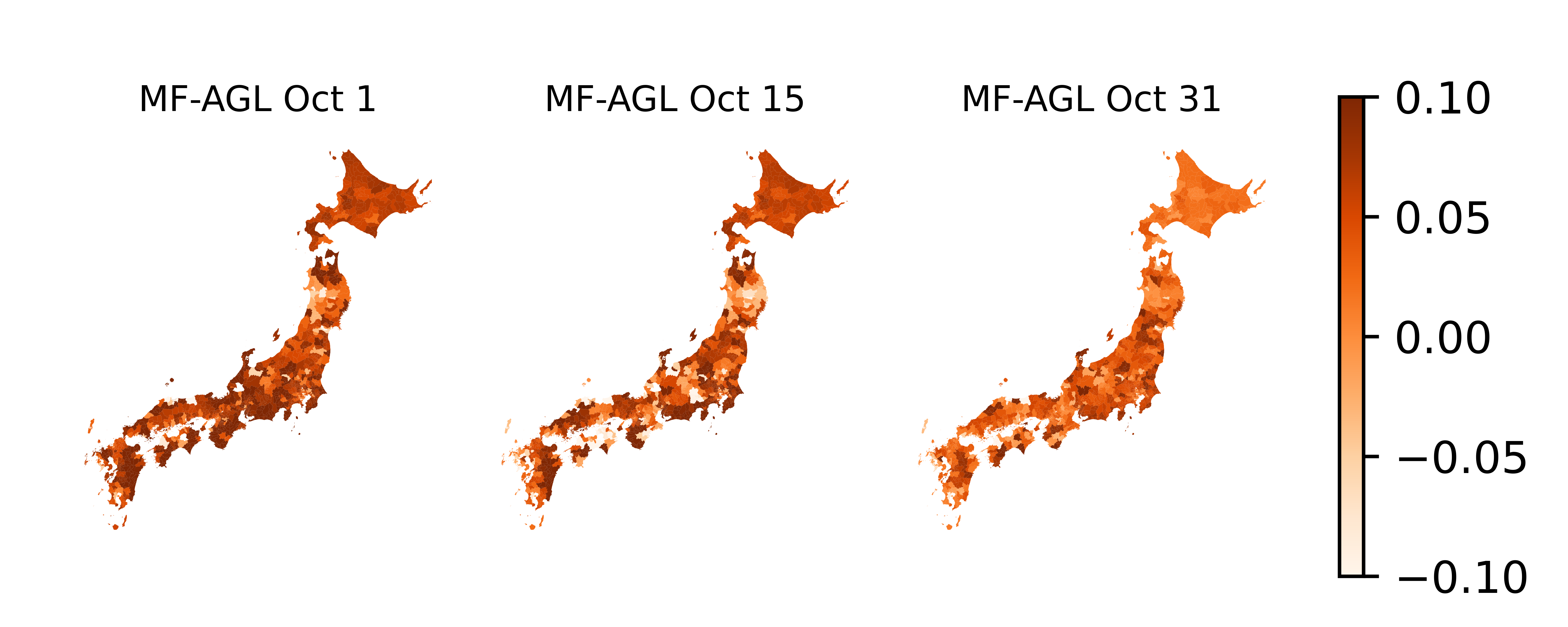}
 \caption{Changes in the number of job applicants in Japan. Due to the limited space of the paper, we exclude Okinawa islands from the map. The color represents good (lighter)-bad(darker) condition of the labor market from the previous year. The prediction is done by MF-AGL model uses the input available at each day.}
 \label{fig:real-time}
\end{figure*}

\begin{figure}[tbhp]
 \centering
 \includegraphics[clip, width = .7\linewidth]{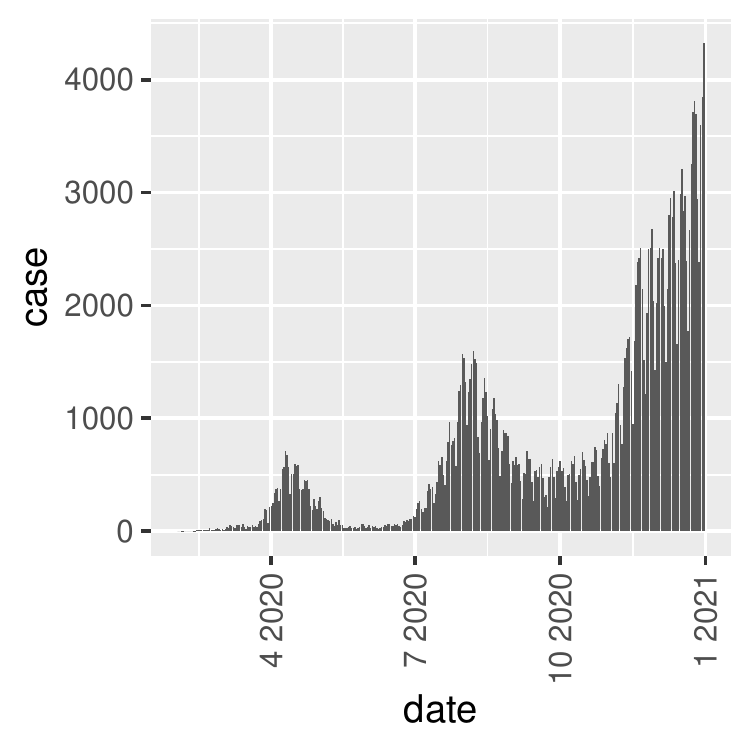}
 \caption{The number of cases (PCR positive) of COVID-19 in Japan.} 
 \label{fig:covid}
\end{figure}

Finally, we demonstrate another practical usefulness of our model. As discussed in Section~\ref{sec:problem setting}, the beauty of our model is real-time updates of forecasting using granular data. Fig.~\ref{fig:real-time} shows how the model updates the prediction using the real-time data. The model shows the year over year for each city and for each day. That is, we first {\it predict} the disaggregate number of job applicants on the same day of the previous year and calculate the year-over-year change in the number of job applicants. The figure implies the rapid improvement in the labor market during October 2020. The possible reason is the peak-out of the second wave of the COVID-19 pandemic. As shown in Fig.~\ref{fig:covid}, the number of cases was dramatically decreased in September and stable in October.

\section{Conclusion}
\label{sec:conclusion}
In this work, we proposed a novel aggregate learning method that adopts to the mixed-frequency data. The model predicts spatio-temporal changes in the economic indices without granular label data. We proposed an LSTM-based simple architecture and the loss function for the training.

We applied the model to a real-world task that is the prediction of the number of job applicants in Japan. Our MF-AGL model well predicts the regional heterogeneity at the sub-prefecture level and also the rapid change in the labor market condition during a month. 

The present model can be applied to broad areas including GDP prediction, labor market prediction, and industrial production prediction. The direction of the future work can be the extension to the other domain and more improvement of the model using more fine-grained architectures.

\begin{acks}

\end{acks}

\clearpage
\bibliographystyle{ACM-Reference-Format}
\bibliography{ref}

\end{document}